\pdfoutput=1

\documentclass[11pt]{article}

\usepackage{acl}

\usepackage{times}
\usepackage{latexsym}

\usepackage[T1]{fontenc}

\usepackage[utf8]{inputenc}

\usepackage{microtype}

\usepackage{amsmath,amssymb}
\usepackage{multirow,booktabs,hhline}
\usepackage{graphicx}
\usepackage{CJKutf8}

\title{QuoteR: A Benchmark of Quote Recommendation for Writing}

\author{
Fanchao Qi$^{1,2}$\thanks{\ \ Equal contribution}\hspace{0.3em},
Yanhui Yang$^{2*}$, 
Jing Yi$^{1,2}$,
Zhili Cheng$^{1,2}$,
\\
{\bf Zhiyuan Liu$^{1,2,3,4}$, Maosong Sun$^{1,2,3,4}$\thanks{\ \  Corresponding author. Email: sms@tsinghua.edu.cn}
}
\\ 
$^{1}$Department of Computer Science and Technology, Tsinghua University\\
Institute for Artificial Intelligence, Tsinghua University, Beijing, China\\
$^{2}$Beijing National Research Center for Information Science and Technology\\
$^{3}$Institute Guo Qiang, Tsinghua University, Beijing, China \\
$^{4}$International Innovation Center of Tsinghua University, Shanghai, China \\
{\tt qfc17@mails.tsinghua.edu.cn, flutter0696@gmail.com}
}

\begin{document}
\maketitle
\begin{abstract}
It is very common to use quotations (quotes) to make our writings more elegant or convincing.
To help people find appropriate quotes  efficiently, the task of quote recommendation is presented, aiming to recommend quotes that fit the current context of writing.
There have been various quote recommendation approaches, but they are evaluated on different unpublished datasets.
To facilitate the research on this task, we build a large and fully open quote recommendation dataset called QuoteR, which comprises three parts including English, standard Chinese and classical Chinese. 
Any part of it is larger than previous unpublished counterparts.
We conduct an extensive evaluation of existing quote recommendation methods on QuoteR.
Furthermore, we propose a new quote recommendation model that significantly outperforms previous methods on all three parts of QuoteR.
All the code and data of this paper can be obtained at \url{https://github.com/thunlp/QuoteR}.

\end{abstract}

\section{Introduction}

A quotation, or quote for short, is a sequence of words that someone else has said or written.
Quotes, especially the famous quotes including proverbs, maxims and other famous sayings, are quite useful in writing --- they can not only help illuminate and emphasize the meaning we want to convey, but also endow our writing with elegance and credibility \citep{cole2008news}.
As a result, the use of quotes is very common and, moreover, universal among all languages. 

\begin{figure}[t!]
\setlength{\belowcaptionskip}{-8pt}   
  \centering
  \includegraphics[width=\linewidth]{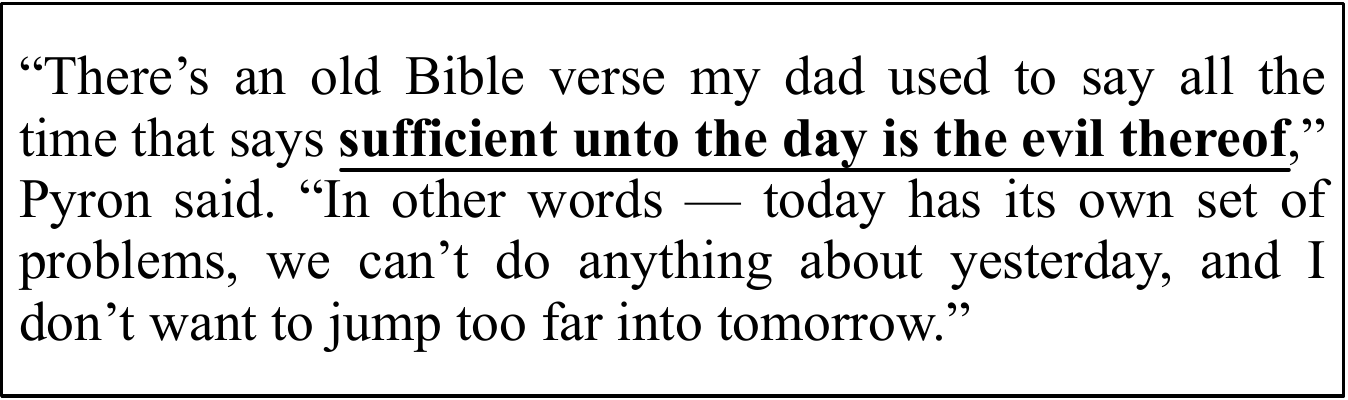} 
  
  \caption{An example of usage of quotes.}
\label{fig:example}
\end{figure}

However, it is not an easy job for ordinary people to promptly come up with appropriate quotes that fit the current context of writing, due to the huge number of quotes.
Search engines can provide some help in finding quotes by keyword matching, but it is often not enough.
Quotes generally express their meanings implicitly by rhetorical devices like metaphor and have different word usages from modern and everyday writing, 
as illustrated in Figure \ref{fig:example}, for which quote search based on keyword matching is ineffective.
In addition, some quote repository websites organize quotes by topic.
However, even after filtering by topic, there are still too many candidate quotes, and selecting a suitable one remains time-consuming.



To tackle these challenges, \citet{tan2015learning} introduce the task of quote recommendation, aiming to automatically recommend suitable quotes given the context of writing.\footnote{This task also has great value to research, as a touchstone for NLP models' abilities in language understanding, semantic matching and linguistic coherence estimation.} 
Afterward, a series of studies propose various approaches to this task \citep{ahn2016quote,tan2016neural,tan2018quoterec}.
However, these studies use different evaluation datasets, and none of them are publicly available. 
The lack of a standard and open dataset is undoubtedly a serious obstacle to the quote recommendation research.

In this paper, to solve this problem, we build a large quote recommendation dataset that is publicly available.
This dataset is named QuoteR (abbreviated from \textbf{Quote} \textbf{R}ecommendataion) and composed of three parts:
(1) the \underline{English} part that comprises 6,108 English quotes 
 with 126,713 contexts; 
(2) the \underline{standard Chinese} (Mandarin) part, which contains 3,004 standard Chinese quotes with 40,842 contexts;
and (3) the \underline{classical Chinese} (Wenyan) part, which comprises 4,438 classical Chinese quotes (including classical poems) and 116,537 contexts.
Any part of this dataset is absolutely larger than, or even doubles, previous closed-source counterparts.

We conduct a fair and extensive evaluation of existing quote recommendation methods on QuoteR with a thorough set of metrics.
By analyzing these methods and their evaluation results, we find two weaknesses of these methods and propose a new method by making corresponding improvements, which we believe would serve as a strong baseline for quote recommendation.

First, most existing methods encode contexts and quotes into vectors for quote-context matching, using LSTM \citep{hochreiter1997long} or CNN \citep{kim2014convolutional} as the encoders.
These encoders have proven inferior to the pre-trained language models like BERT \citep{devlin2019bert}, which limits the final quote recommendation performance.
Therefore, we try to utilize a pre-trained language model, specifically BERT, as the sentence encoders to learn representations of quotes and contexts.
Considering the huge compute resulting from the large scale of the dataset and the BERT model, it is nontrivial to train the context and quote encoders simultaneously.
We design an ingenious training strategy to address this issue. 

Second, it is harder to learn good representations for quotes compared with contexts, because most quotes are quite pithy, and their words usually carry rich semantics, as shown in Figure \ref{fig:example}.
Existing methods, however, do not address this challenge well.
To handle this challenge, we incorporate a kind of general lexical knowledge, namely \textit{sememes}, into the quote encoder, aiming to improve the representations of quotes.
A sememe is defined as the minimum semantic unit in linguistics \citep{bloomfield1926set}, and the sememes of a word atomically interpret the meaning of the word.
Incorporating sememes can bring more semantic information for quote representation learning and conduce to a better quote vector. 

In experiments, we demonstrate that both the utilization of BERT and the incorporation of sememes substantially improve quote recommendation performance. 
And the sememe-incorporated BERT-based model significantly outperforms all previous methods on QuoteR. 
Moroever, ablation and case studies as well as human evaluation further prove its effectiveness.



To conclude, our contributions are threefold: (1) building a large and the first open quote recommendation dataset; (2) conducting an extensive and fair evaluation of existing quote recommendation methods; (3) proposing a quote recommendation model that outperforms all previous methods and can serve as a strong baseline for future research. 

\section{Related Work}

\subsection{Quote Recommendation}
\label{sec:qr_work}
The task of quote recommendation is originally presented in \citet{tan2015learning}.
They propose a learning-to-rank framework for this task, which integrates 16 hand-crafted features. 
\citet{tan2016neural} and \citet{tan2018quoterec} introduce neural networks to the quote recommendation task.
They use LSTMs to learn distributed vector representations of contexts and quotes and conduct sentence matching with these vectors.
\citet{ahn2016quote} combine four different quote recommendation approaches including matching granularity adjustment (a statistical context-quote relevance prediction method), random forest, CNN and LSTM.

In addition quote recommendation for writing, some studies focus on recommending quotes in dialog.
\citet{lee2016quote} propose an LSTM-CNN combination model to recommend quotes according to Twitter dialog threads, i.e., sequences of linked tweets.
\citet{wang2020continuity} utilize an encoder-decoder framework to generate quotes as response, based on the separate modeling of the dialog history and current query.
\citet{wang2021quotation} adopt a semantic matching fashion, which encodes the multi-turn dialog history with Transformer \citep{vaswani2017attention} and GRU \citep{cho2014learning} and encodes the quote with Transformer.

In terms of the datasets of quote recommendation for writing, \citet{tan2015learning} construct an English dataset comprising 3,158 quotes and 64,323 contexts extracted from e-books in Project Gutenberg.\footnote{\url{https://www.gutenberg.org/}}
\citet{ahn2016quote} build a similar English dataset that contains 400 most frequent quotes with contexts from e-books in Project Gutenberg and blogs.
\citet{tan2018quoterec} build a classical Chinese poetry quotation dataset that comprises over 9,000 poem sentences with 56,949 contexts extracted from Chinese e-books on the Internet.
Unfortunately, all these datasets are not publicly available. 

\subsection{Content-based Recommendation}
Quote recommendation is essentially a kind of content-based recommendation task \citep{pazzani2007content}, which is aimed at recommending products to users according to product descriptions and users' profiles. 

A closely related and widely studied task is content-based citation recommendation \citep{strohman2007recommending}, especially local citation recommendation that recommends related papers given a particular context of academic writing \citep{he2010context,huang2012recommending,huang2015neural}.
Compared with quote recommendation, this task is targeted at structured documents (papers), which are much longer and possess abundant information such as title, abstract and citation relations that are useful for recommendation.
Quotes are shorter and usually have no available information except the text, which renders quote recommendation more challenging.

Another highly related but niche task is idiom recommendation \citep{liu2018modelling,liu2019neural}, which aims to recommend appropriate idioms for a given context.
Existing idiom recommendation methods are essentially covered by the quote recommendation methods described in §\ref{sec:qr_work}.
\citet{liu2018modelling} recommend idioms by learning representations of the contexts and idioms, similar to the context-quote relevance-based quote recommendation methods \citep{ahn2016quote,tan2018quoterec}.
The difference lies in the use of word embeddings of idioms rather than a sentence encoder. 
\citet{liu2019neural} regard idiom recommendation as a context-to-idiom machine translation problem and use an LSTM-based encoder-decoder framework, which is similar to \citet{wang2020continuity}.


\subsection{Other Quote-related Tasks}
In addition to quote recommendation, there are some other quote-related tasks.
For example, quote detection (or recognition) that is aimed at locating spans of quotes in text \citep{pouliquen2007automatic,scheible2016model,pareti2013automatically,papay2019quotation}, and quote attribution that intends to automatically attribute quotes to speakers in the text  \citep{elson2010automatic,o2012sequence,almeida2014joint,muzny2017two}.
Different from quote recommendation that focuses on famous quotes, these tasks mainly deal with the general quotes of utterance.

\section{Task Formulation}
\label{sec:formalize}

Before describing our dataset and model, we first formulate the task of quote recommendation for writing and introduce several basic concepts, 
most of which follow previous work \citep{tan2015learning}. 

For a piece of text containing a quote $q$, the text segment occurring before the quote is named \textit{left context} $c_l$ while the text segment occurring after the quote is named \textit{right context} $c_r$.
The concatenation of left and right contexts form the \textit{quote context} $c=[c_l;c_r]$.
Suppose there is a \textit{quote set} that comprises all the known candidate quotes $Q=\{q_1, \cdots, q_{|Q|}\}$, where $|\cdot|$ denotes the cardinality of a set.  

In the task of quote recommendation for writing, a \textit{query context}  $c=[c_l;c_r]$ is given, and the \textit{gold quote} $q_c$ is wanted, where the query context is the context provided by the user and the gold quote is the quote in the quote set that fits the query context best.
Theoretically, a query context may have more than one gold quote because there are some quotes that convey almost the same meaning.
Following previous work \citep{tan2015learning,ahn2016quote}, for simplicity, we only regard the quote that actually appears together with the query context in corpora as the gold quote. 

For a quote recommendation model, given the quote set $Q$, its input is a query context $c=[c_l;c_r]$, and it is supposed to calculate a rank score for each candidate quote in $Q$ and output a quote list according to the descending rank scores.

\section{Dataset Construction}

In this section, we present the building process and details of the QuoteR dataset.

\subsection{The English Part}

We begin with the English part. 
We choose the popular and free quote repository  website Wikiquote\footnote{\url{https://en.wikiquote.org}} as the source of English quotes.
We download its official dump 
and extract over 60,000 English quotes in total to form the quote set. 
We notice that previous work \citep{tan2015learning,ahn2016quote} collects quotes from another website named Library of Quotes, but this website has closed down.


To obtain real contexts of quotes, we use three corpora.
The first is the Project Gutenberg corpus that previous studies use, which comprises over 50,000 e-books.
The second corpus is BookCorpus containing about 11,000 e-books \citep{zhu2015aligning}.
In addition to the two book corpora, we 
use the OpenWebText corpus \citep{Gokaslan2019OpenWeb} 
which is composed of text from web pages and has different text styles from books. 
The total size of the raw text of the three corpora reaches 48.8 GB.

We search all the corpora for the occurrences of quotes in the quote set.
Some quotes are composed of multiple sentences,  and only part of them are cited in some cases.
To cope with this situation, we split each quote into sentences using Stanza \citep{qi2020stanza} and then search for each constituent sentence in the corpora. 
If multiple constituent sentences of a quote appear sequentially, we combine them into an occurrence of the quote.
Compared with previous work that searches for quotes as a whole \citep{tan2015learning,ahn2016quote}, we can find more quote occurrences.

For each quote occurrence, we take the 40 words 
preceding and following it as its left and right contexts, respectively. 
The concatenation of the left and right contexts forms a context, and a context and the corresponding quote form a context-quote pair.
We remove the repeated context-quote pairs and filter out the quotes appearing less than 5 times in the corpora.
To avoid dataset imbalance, we randomly select 200 context-quote pairs for a quote appearing more than 200 times and discard its other context-quote pairs.
Finally, we obtain 126,713 context-quote pairs involving 6,108 different quotes, which form the English part of QuoteR.

We split all the context-quote pairs into training, validation and test sets roughly in the ratio 8:1:1, making sure that all the quotes appear in the validation and test sets while 100 quotes do not appear in the training set. 
We split the dataset in this way in order to observe how quote recommendation models perform in the zero-shot situation, where the model has never seen the gold quote of some validation/test contexts during training.
The statistics of the final split dataset are listed in Table \ref{tab:stat-dataset}.

\begin{table}[!t]
\centering
\resizebox{1.03\linewidth}{!}{
    \begin{tabular}{crrrr}
    \toprule
  {Part}   & \multicolumn{1}{c}{Train}  & \multicolumn{1}{c}{Validation}  & \multicolumn{1}{c}{Test} & \multicolumn{1}{c}{Total} \\
    \midrule
  English & 101,171/6,008 & 12,771/6,108 & 12,771/6,108 & 126,713/6,108 \\
  sChinese & 32,472/2,904 & 4,185/3,004 & 4,185/3,004 & 40,842/3,004 \\
  cChinese & 93,031/4,338 & 11,753/4,438 & 11,753/4,438 & 116,537/4,438\\
    \bottomrule
    \end{tabular}
}
\caption{Statistics of the three parts of QuoteR. sChinese and cChinese refer to standard and classical Chinese, respectively. Each item like $m$/$n$ means $m$ context-quote pairs involving $n$ quotes. Appendix \ref{sec:stat} gives more detailed statistics.}
\label{tab:stat-dataset}
\end{table}

\subsection{The Standard Chinese Part}
We gather standard Chinese quotes from a large quote collection website named Juzimi\footnote{\url{https://www.juzimi.com/}}.
More than 32,000 standard Chinese quotes are collected altogether.
To obtain quote contexts, we use two corpora including a corpus composed of answer text from a Chinese QA website\footnote{\url{https://github.com/brightmart/nlp_chinese_corpus}} and a large-scale book corpus that we specifically build and comprises over 8,000 free Chinese e-books.
The total size of the two corpora is about 32 GB. 

Then we use the same method in building the English part to extract quote occurrences from the corpora.
Since Chinese is not naturally word-segmented, we take the 50 characters (rather than words)
before and after a quote occurrence as the left and right contexts.
In addition, since there are fewer quotes and contexts for the standard Chinese part, we reduce the minimum number of occurrences for a selected quote to 3, and the maximum number of retained contexts per quote to 150. 
After deduplication and filtering, we obtain the standard Chinese part of QuoteR, which has 40,842 context-quote pairs involving 3,004 quotes.

We split the standard Chinese part in the same way as the English part, and the statistics are also shown in Table \ref{tab:stat-dataset}.

\subsection{The Classical Chinese Part}
Classical Chinese quotes, including classical poems and proverbs, are often cited in standard Chinese writing.
Considering that classical Chinese is very different from standard Chinese, we separate classical Chinese quotes from standard Chinese ones. 
We collect over 17,000 classical Chinese quotes from Gushiwenwang,\footnote{\url{https://www.gushiwen.org/}} a classical Chinese poetry and literature repository website, and aforementioned Juzimi.\footnote{Juzimi provides the dates when the quotes appear so that we can distinguish classical and standard Chinese quotes.}

Then we adopt the same way as standard Chinese to extract context-quote pairs from the two Chinese corpora and conduct deduplication and filtering. 
Finally, we obtain the classical Chinese part of QuoteR that comprises 116,537 context-quote pairs of 4,438 quotes.
The statistics of this part after splitting are also in Table \ref{tab:stat-dataset}.

\subsection{Quality Assessment by Human}
After the construction of QuoteR, we assess its quality by human.
For each part, we randomly sample 100 context-quote pairs, and ask three annotators to independently determine whether each quote fits the corresponding context.
The final results are obtained by voting.
Finally, 99/98/94 context-quote pairs are regard as suitable for the three parts, respectively.
The results verify the quality of QuoteR, which is expected because the data are extracted from high-quality corpora like books.

\section{Methodology} 
In this section, we elaborate on our proposed quote recommendation model.
This model is based on the representative pre-trained language model BERT \citep{devlin2019bert}, but can be readily adapted to other pre-trained language models. 


\subsection{Basic Framework}
\label{sec:framework}
Similar to most previous methods \citep{tan2016neural,ahn2016quote}, we use BERT as the text encoder to learn vector representations of contexts and quotes, and then calculate the similarity between the representations of the query context and a candidate quote as the rank score of the quote.



\subsubsection*{Learning Representations of Quotes}
We first obtain the representations of quotes.
Formally, for a candidate quote comprising $m$ tokens $q=\{x_1,\cdots,x_m\} \in Q$, we feed it into BERT and obtain a series of hidden states:
\begin{equation}
\fontsize{9.5pt}{12pt}
\selectfont
  \mathbf{h}^{q}_{\texttt{[C]}},\mathbf{h}^{q}_{1},\cdots,\mathbf{h}^{q}_{m}={\rm BERT}^{q}(\texttt{[C]},x_1,\cdots,x_m),
\end{equation}
where \texttt{[C]} denotes the special \texttt{[CLS]} token in BERT that is added to the front of a sequence.
Following \citet{devlin2019bert}, we use the hidden state of \texttt{[C]} as the representation of the quote: $\mathbf{q}=\mathbf{h}^{q}_{\texttt{[C]}}$.
The representations of all quotes 
form the quote representation matrix $\mathbf{Q}=[\mathbf{q}_1,\cdots,\mathbf{q}_{|Q|}$].

\subsubsection*{Learning Representations of Contexts}

We can use another BERT as the context encoder to obtain the representation of the query context $c=[c_l;c_r]$.
Considering the context is composed of left and right contexts that are not naturally joined, we can insert an additional separator token between them before feeding them into BERT:
\begin{equation}
\fontsize{10.5pt}{12pt}
\selectfont
  \mathbf{h}^{c}_{\texttt{[C]}},\cdots={\rm BERT}^{c}(\texttt{[C]},c_l,\texttt{[S]},c_r),
\end{equation}  
where \texttt{[S]} is the sentence separator token \texttt{[SEP]} in BERT.
We can also use the hidden state of \texttt{[C]} as the representation of the context: $\mathbf{c}=\mathbf{h}^c_{\texttt{[C]}}$.

However, it is actually inconsistent with the general use of BERT. 
Whether in pre-training or fine-tuning, when the input to BERT is two text segments connected by the separator token, the hidden state of \texttt{[CLS]} is only used to classify the relation between the two segments, e.g., to predict whether the second segment is the actual next sentence of the first segment in the next sentence prediction (NSP) pre-training task \citep{devlin2019bert}.

We turn to another pre-training task of BERT, masked language modeling (MLM), which is a cloze task \citep{taylor1953cloze} aimed at predicting masked tokens.
Specifically, some tokens in a text sequence are randomly substituted by the special \texttt{[MASK]} tokens and the hidden states of the \texttt{[MASK]} tokens are fed into a classifier to predict the original tokens.
Quote recommendation given context can be regarded as a special cloze task whose object of prediction is quotes rather than tokens.
Inspired by the MLM pre-training task, we propose another way to learn the context representation by inserting an additional \texttt{[MASK]} token:
\begin{equation}
\fontsize{9.5pt}{12pt}
\selectfont
  \mathbf{h}^{c}_{\texttt{[C]}},\cdots,\mathbf{h}^{c}_{\texttt{[M]}},\cdots={\rm BERT}^{c}(\texttt{[C]},c_l,\texttt{[M]},c_r),
\end{equation} 
where \texttt{[M]} is the \texttt{[MASK]} token.
We use the hidden state of \texttt{[M]} as the representation of the query context: $\mathbf{c}=\mathbf{h}^c_{\texttt{[M]}}$.\footnote{The hidden state of \texttt{[M]} can also be regarded as the representation of the \textit{required} quote for the query context. In this view, the rank score in Eq. \eqref{eq:rank_score} is actually calculated by the similarity between a candidate quote and the required quote.} 


\subsubsection*{Calculating Rank Scores of Candidate Quotes}
After obtaining the representations of all candidate quotes and the query context, the rank score of a candidate quote can be calculated by softmax:
\begin{equation}
\label{eq:rank_score}
  \mathbf{p}={\rm softmax}(\mathbf{Q}^\top\mathbf{c}),
\end{equation}
where $\mathbf{p}$ is a normalized probability vector whose $i$-th element is the rank score of the $i$-th quote. 


\subsection{Training Strategy}
\label{sec:train}
As in previous work \citep{tan2016neural}, we can simply use the cross-entropy loss to train the quote and context encoders simultaneously. 
However, there are two problems.
(1) For each context in the training set, the quote encoder needs to be updated for every quote in the quote set.
In other words, the BERT-based quote encoder would be fine-tuned thousands of times per training instance, which requires formidably big GPU memory and long training time.\footnote{We find that four 16-GB GPUs would be out of memory during training even though we set the batch size to 1.} 
(2) The huge imbalance between positive and negative samples (one vs. several thousands) would weaken the capacity of the quote encoder and, in turn, impair the final quote recommendation performance.

A simple solution is to freeze the quote encoder during training, i.e., use the raw pre-trained BERT as the quote encoder, and train the context encoder only.
But the untrained quote encoder would decrease final quote recommendation performance, as demonstrated in later experiments. 
To address these issues, inspired by the study on noise contrastive estimation (NCE) \citep{gutmann2012noise}, 
we adopt the negative sampling strategy in training.
For each context-quote pair, we select some non-gold quotes as negative samples, and calculate a pseudo-rank score of the gold quote among the selected quotes.
Formally, for a context-quote pair $(c,q)$, the pseudo-rank score of $q$ is
\begin{equation}
\fontsize{10.5pt}{12pt}
\selectfont
p^*=\frac{e^{\mathbf{q}\cdot\mathbf{c}}}{e^{\mathbf{q}\cdot\mathbf{c}}+\sum_{q^*\in \mathbb{N}(q)}e^{\mathbf{q^*}\cdot\mathbf{c}}},
\end{equation}
where $\mathbb{N}(q)$ is the set of quotes selected as negative samples.
Then the training loss is the cross-entropy based on the pseudo-rank score: $\mathcal{L}=-\log(p^*)$.


The problem about quote encoder training has been largely solved, but the context encoder may be under-trained.
The context encoder needs to process lots of contexts and thus requires more training than the quote encoder.
Therefore, we adopt a two-stage training strategy.
After the simultaneous training of quote and context encoders in the first stage, we continue to train the context encoder while freezing the quote encoder in the second stage.
The training loss of the second stage is the cross-entropy loss among all quotes. 

\subsection{Incorporation of Sememes}
Most quotes are quite pithy, and thus it is usually hard to learn their representations well.
To obtain better quote representations, previous work tries incorporating external information, including the topic and author information of quotes, in the quote encoder \citep{tan2016neural,tan2018quoterec}.
Although helpful, this external information is not always available or accurate --- quite a few quotes are anonymous, and the topics attributed to quotes are usually from crowdsourcing and uninspected. 

We propose to incorporate sememe knowledge into quote representation learning, which is more general (every word can be annotated with sememes) and credible (the sememe annotations of words are given by experts).
A sememe is the minimum semantic unit of human languages \citep{bloomfield1926set}, and it is believed that meanings of all words can be represented by a limited set of sememes. 
Sememe knowledge bases like HowNet \citep{dong2006hownet} use a set of predefined sememes to annotate words, so that the meaning of a word can be precisely expressed by its sememes.
With the help of such sememe knowledge bases, sememes have been successfully utilized in various NLP tasks \citep{qi2021sememe}, including semantic composition \citep{qi2019modeling}, word sense disambiguation \citep{hou2020try}, reverse dictionary \citep{zhang2020multi}, adversarial attacks \citep{zang2020word}, backdoor learning \citep{qi2021turn}, etc.

Inspired by the studies on incorporating sememes into recurrent neural networks \citep{qin2020improving} and transformers \citep{zhang2020enhancing} to improve their representation learning ability, we adopt a similar way to incorporate sememes into the quote encoder.
We simply add the average embedding of a word's sememes to every token embedding of the word in BERT.
Formally, for a word in a quote that is divided into $n$ tokens after tokenization $w={x_1,\cdots,x_n}$, the embedding of its each token $\mathbf{x}_i$ is transformed into

\begin{equation}
\fontsize{10.5pt}{12pt}
\selectfont
  \mathbf{x}_i\rightarrow \mathbf{x}_i+\frac{\alpha}{|\mathbb{S}(w)|}\sum_{s_j\in\mathbb{S}(w)}\mathbf{s}_j, \  \forall i=1,\cdots,n
\end{equation}
where $\mathbb{S}(w)$ is the sememe set of the word $w$, and $\alpha$ is a hyper-parameter controlling the weight of sememe embeddings.
Following previous work \citep{qin2020improving}, the sememe embeddings are randomly initialized and updated during training. 

\begin{table*}
\centering

\resizebox{1.0\textwidth}{!}{%
\begin{tabular}{@{}l|cccc|ccrc|ccrc@{}}
\toprule
\multicolumn{1}{c|}{Part} & \multicolumn{4}{c|}{English} & \multicolumn{4}{c|}{Standard Chinese} & \multicolumn{4}{c}{Classical Chinese} \\ \midrule
\multicolumn{1}{c|}{Model} & MRR & NDCG &  \multicolumn{1}{c}{$\tilde{R}$ /$\bar{R}$ /$\sigma_R$} & Recall@1/10/100 & MRR & NDCG &   \multicolumn{1}{c}{$\tilde{R}$ /$\bar{R}$ /$\sigma_R$} & Recall@1/10/100 & MRR & NDCG &   \multicolumn{1}{c}{$\tilde{R}$ /$\bar{R}$ /$\sigma_R$} & Recall@1/10/100 \\ \midrule
CRM & 0.192 & 0.193 & 599/1169/1408 & 16.51/23.66/32.78 & 0.397 & 0.407 & 13/325/584 & 33.60/49.32/61.70 & 0.198 & 0.203 & 166/548/811 & 14.52/28.79/44.51 \\
LSTM & 0.321 & 0.320 & 30/334/727 & 27.23/40.78/62.47 & 0.292 & 0.290 & 48/338/574 & 24.78/37.71/58.06 & 0.247 & 0.245 & 56/341/633 & 20.08/33.23/56.96 \\
top-k RM & 0.422 & 0.431 & 6/548/1243 & 35.99/53.31/66.20 & 0.480 & 0.494 & 3/377/774 & 40.17/60.67/72.26 & 0.294 & 0.299 & 48/511/980 & 23.54/39.58/56.90 \\
NNQR & 0.318 & 0.319 & 31/359/773 & 26.78/41.10/61.29 & 0.271 & 0.271 & 54/348/595 & 22.94/35.72/57.18 & 0.272 & 0.270 & 41/310/620 & 22.03/36.59/60.63 \\
N-QRM & 0.365 & 0.368 & 28/777/1465 & 32.24/44.41/58.26 & 0.343  & 0.347 & 55/575/890 & 30.20/41.22/54.15 & 0.287 & 0.288 & 98/917/1373 & 24.88/35.02/49.49 \\
Transform & 0.561 & 0.568 & \underline{1}/241/749 & 50.11/65.88/79.98 & 0.512 & 0.519 & \underline{2}/271/576 & 45.50/60.31/72.83 & 0.449 & 0.453 & 5/269/663 & 39.01/55.78/73.58 \\
BERT-Sim & 0.526 & 0.529 & 2/487/1064 & 49.38/58.05/67.75 & 0.500 & 0.508 & \underline{2}/229/511 & 44.47/59.07/72.21 & 0.439 & 0.443 & 7/320/711 & 38.85/53.04/68.32 \\
BERT-Cls & 0.310 & 0.329 & 7/134/453 & 18.15/57.11/82.05 & 0.378 & 0.395 & 5/152/413 & 26.88/57.90/78.38 & 0.330 & 0.345 & 8/\textbf{135}/\textbf{377} & 21.93/54.27/78.75 \\ \midrule
Ours & \textbf{0.572} & \textbf{0.580} & \underline{1}/\textbf{123}/\textbf{433} & {50.74}/\textbf{69.03}/\textbf{83.84} & \textbf{0.541} & \textbf{0.548} & \underline{2}/\textbf{139}/\textbf{370} & \textbf{47.91}/\textbf{64.97}/\textbf{79.35} & \textbf{0.484} & \textbf{0.490} & \underline{3}/{146}/{422} & \textbf{41.67}/\textbf{60.78}/\textbf{79.38} \\
\multicolumn{1}{r|}{-Sememe} & 0.568 & 0.574 & \underline{1}/145/492 & \textbf{51.05}/67.07/82.34 & 0.535 & 0.543 & \underline{2}/160/402 & 47.62/63.66/77.68 & 0.475 & 0.481 & \underline{3}/152/435 & 40.93/60.26/78.39 \\
\multicolumn{1}{r|}{-ReTrain} & 0.299 & 0.307 & 12/176/503 & 20.46/47.89/75.74 & 0.255 & 0.260 & 20/210/435 & 16.87/42.94/68.43 & 0.265 & 0.269 & 17/184/450 & 17.87/43.56/72.89 \\
\multicolumn{1}{r|}{-SimTrain} & 0.529 & 0.532 & 2/467/1060 & 49.31/58.97/69.48 & 0.519 & 0.526 & \underline{2}/204/489 & 46.00/62.03/75.34 & 0.465 & 0.470 & 4/310/713 & 41.40/55.53/70.09 \\ \bottomrule
\end{tabular}%
}
\caption{Quote recommendation results on the three parts of QuoteR. Recall@1/10/10 is percentage. The \textbf{boldfaced} results exhibit statistically significant improvement over the other results with \textit{p}$<$0.1 given by paired \textit{t}-tests, and the \underline{underlined} results mean no significant difference. The same is true for the following Tables.}
\label{tab:main}
\end{table*}

\section{Experiments}
In this section, we evaluate our model and previous quote recommendation methods on QuoteR.

\subsection{Approaches for Comparison}
We have three groups of approaches for comparison.
The first group consists of two methods that widely serve as baselines in previous studies.
(1.1) \textbf{CRM}, namely context-aware relevance model \citep{he2010context} that recommends the quote whose known contexts are most similar to the query context.
(1.2) \textbf{LSTM}, which uses two LSTM encoders to learn representations of quotes and contexts. 

The second group includes representative approaches proposed in previous studies. 
(2.1) \textbf{top-k RM}, namely top-k rank multiplication \citep{ahn2016quote}, which is a rank aggregation method based on the ensemble of a statistical method, random forest, CNN and LSTM.
(2.2) \textbf{NNQR} \citep{tan2016neural}, which reforms LSTM by incorporating additional quote information (topic and author) into the quote encoder and perturbing the word embeddings of quotes.
(2.3) \textbf{N-QRM} \citep{tan2018quoterec}, which further improves NNQR mostly by adjusting the training loss to prevent overfitting. 
(2.4) \textbf{Transform} \citep{wang2021quotation}, which uses Transformer+GRU to encode contexts and transforms context embeddings into the space of quote embeddings learned from another Transformer.\footnote{It is originally designed to recommend quotes in dialog, and we adapt it to the writing situation. It is also the only adaptable method of other content-based recommend tasks.}


The third group comprises two BERT-based approaches that are frequently utilized in 
sentence matching and sentence pair classification.
(3.1) \textbf{BERT-Sim}, which is the vanilla BERT-based model discussed in §\ref{sec:framework}.
It directly uses the hidden states of the \texttt{[CLS]} tokens as the representations of both quotes and contexts, and freezes the quote encoder during training, as explained in §\ref{sec:train}.
(3.2) \textbf{BERT-Cls}, which conducts a binary classification for the concatenation of the query context and a candidate quote. 

\subsection{Evaluation Metrics}
Following previous work \citep{ahn2016quote,tan2018quoterec}, we use three evaluation metrics:
(1) Mean reciprocal rank (\textbf{MRR}), the average reciprocal values of the ranks of the gold quotes; 
(2) Normalized discounted cumulative gain (\textbf{NDCG@K}) \citep{jarvelin2002cumulated}, a widely used measure of ranking quality  and is computed by
  \begin{equation} 
  {\rm NDCG@K}=Z_{K}\sum_{i=1}^{K}\frac{2^{r(i)}-1}{\log_2(i+1)},
  \end{equation}
  where $r(i)=1$ if the $i$-th quote is the gold quote, otherwise $r(i)=0$, $Z_{K}=1$ is a normalization constant. We report the average of NDCG@5 scores of all the evaluated query contexts.
(3) \textbf{Recall@K}, the proportion of query contexts whose gold quotes are ranked in respective top $K$ candidate quotes, $K=\{1,10,100\}$.

Besides, we use another three evaluation metrics: 
(4) \textbf{Median Rank} ($\tilde{R}$), (5) \textbf{Mean Rank} ($\bar{R}$) and (6) \textbf{Rank Variance} ($\sigma_R$), the median, average and standard deviation of the ranks of gold quotes.

The higher MRR, NDCG@K and Recall@K and the lower $\tilde{R}$, $\bar{R}$ and $\sigma_R$ are, the better a model is.

\begin{table*}[t!]
\centering

\resizebox{1.0\textwidth}{!}{%
\begin{tabular}{@{}l|cccc|ccrc|ccrr@{}}
\toprule
\multicolumn{1}{c|}{Part} & \multicolumn{4}{c|}{English} & \multicolumn{4}{c|}{Standard Chinese} & \multicolumn{4}{c}{Classical Chinese} \\ \midrule
\multicolumn{1}{c|}{Model} & MRR & NDCG &  \multicolumn{1}{c}{$\tilde{R}$ /$\bar{R}$ /$\sigma_R$} & Recall@1/10/100 & MRR & NDCG &   \multicolumn{1}{c}{$\tilde{R}$ /$\bar{R}$ /$\sigma_R$} & Recall@1/10/100 & MRR & NDCG &   \multicolumn{1}{c}{$\tilde{R}$ /$\bar{R}$ /$\sigma_R$} & Recall@1/10/100 \\ \midrule
CRM & 0.154 & 0.156 & 353/948/1297 & 11.88/21.78/33.66 & 0.292 & 0.296 & 124/401/524 & 25.28/35.39/48.43 & 0.141 & 0.146 & 276/587/763 & {9.88/19.75/34.57} \\
LSTM & 0.272 & 0.271 & 89/552/992 & 23.38/33.87/51.12 & 0.210 & 0.208 & 146/483/662 & 18.26/27.67/45.50 & 0.182 & 0.178 & 117/465/750 & 13.87/25.44/47.80 \\
top-k RM & 0.360 & 0.366 & 30/833/1497 & 31.20/44.55/56.80 & 0.350 & 0.358 & 38/620/926 & 29.77/44.40/55.53 & 0.276 & 0.280 & 77/645/1088 & 22.61/36.16/52.57 \\
NNQR & 0.267 & 0.266 & 98/592/1043 & 22.82/33.48/50.28 & 0.224 & 0.223 & 145/495/683 & 17.16/27.67/45.81 & 0.189 & 0.187 & 98/441/766 & 14.18/26.86/50.29 \\
N-QRM & 0.270 & 0.272 & 156/1145/1735 & 23.40/33.18/46.54 & 0.266  & 0.270 & 287/778/946 & 21.27/30.63/42.32 & 0.215 & 0.215 & 356/1232/1505 & 17.72/27.13/40.73 \\
Transform & 0.438 & 0.443 & 6/429/1036 & 38.47/53.43/68.65 & 0.371 & 0.374 & 29/465/748 & 32.54/44.83/58.04 & 0.331 & 0.334 & 29/435/842 & 27.76/42.87/60.85 \\
BERT-Sim & 0.399 & 0.401 & 44/839/1407 & 36.95/44.75/54.32 & 0.364 & 0.370 & 41/431/695 & 31.71/44.28/56.18 & 0.310 & 0.313 & 56/522/902 & 26.32/39.05/54.56 \\
BERT-Cls & 0.265 & 0.275 & 15/\textbf{237}/\textbf{640} & 16.75/45.37/71.77 & 0.213 & 0.220 & 24/318/646 & 12.47/40.53/64.67 & 0.204 & 0.208 & 25/253/568 & 11.50/38.27/66.73 \\ 
\midrule
Ours & \textbf{0.456} & \textbf{0.462} & \textbf{4}/254/685 & \textbf{39.62}/\textbf{56.21}/\textbf{73.26} & \textbf{0.413} & \textbf{0.419} & \textbf{7}/\textbf{97}/\textbf{186} & \textbf{34.64}/\textbf{53.29}/\textbf{75.91} & \textbf{0.409} & \textbf{0.411} & \textbf{9}/\textbf{196}/\textbf{419} & \textbf{35.22}/\textbf{51.47}/\textbf{70.82} \\
\bottomrule
\end{tabular}%
}
\caption{Quote recommendation results on the three parts of QuoteR, \textbf{given the left context only}.}
\label{tab:left}
\end{table*}

\subsection{Implementation Details}
We use BERT$_{\text{BASE}}$ for both English and Chinese from Transformers \citep{wolf2020transformers}.
We use the AdamW optimizer \citep{loshchilov2018fixing} with an initial learning rate 5e-5 that gradually declines to train our model.
We randomly select $N$ negative samples, and $N$ is tuned in \{4,9,\underline{19},29,39\} on the validation set.
The weight of sememe embeddings $\alpha$ is tuned in \{0.1, 0.2, \underline{0.5}, 1.0, 2.0\}.
The underlined numbers are final picks.
For the previous methods, we use their original hyperparameters and experimental settings given in the papers.

\subsection{Main Results}
Table \ref{tab:main} lists the evaluation results of different methods on the three parts of QuoteR. 
We observe that (1) our method achieves the best overall results and displays its superiority to other methods; 
(2) the two BERT-based models, especially BERT-Sim, yield quite high performance, which reflects the importance of a powerful sentence encoder to quote recommendation; 
(3) among the three parts, almost all methods perform worse on Classical Chinese, which is presumably because Chinese BERT is pre-trained on standard Chinese corpora and not suitable to encode the classical Chinese quotes.


\subsubsection*{Ablation Study}
We conduct ablation studies to investigate the effectiveness of our training strategy and the incorporation of sememes.
We first remove the incorporation of sememes (-Sememe), then further do not separately train the context encoder after the simultaneous training of the context and quote encoders (-ReTrain), and finally discard the simultaneous training of the two encoders and train the context encoder only (-SimTrain). 
-SimTrain differs BERT-Sim only in the choice of context representation (\texttt{[MASK]} vs. \texttt{[CLS]}).

The results of ablation studies are given in the last three rows of Table \ref{tab:main}.
We have the following observations:
(1) -Sememe causes consistent performance decline as compared to Ours, which demonstrates the role of sememes in improving quote encoding, thereby benefiting quote recommendation;
(2) the performance of -ReTrain is pretty poor, which reflects the necessity of separate training for the context encoder after simultaneous training;
(3) -SimTrain is inferior to -Sememe, which displays the usefulness of simultaneously training the two encoders;
(4) -SimTrain outperforms BERT-Sim, proving the superiority of choosing \texttt{[MASK]} to represent contexts in our method.

%

\begin{figure}[t!] 
  \centering
  \includegraphics[width=\linewidth]{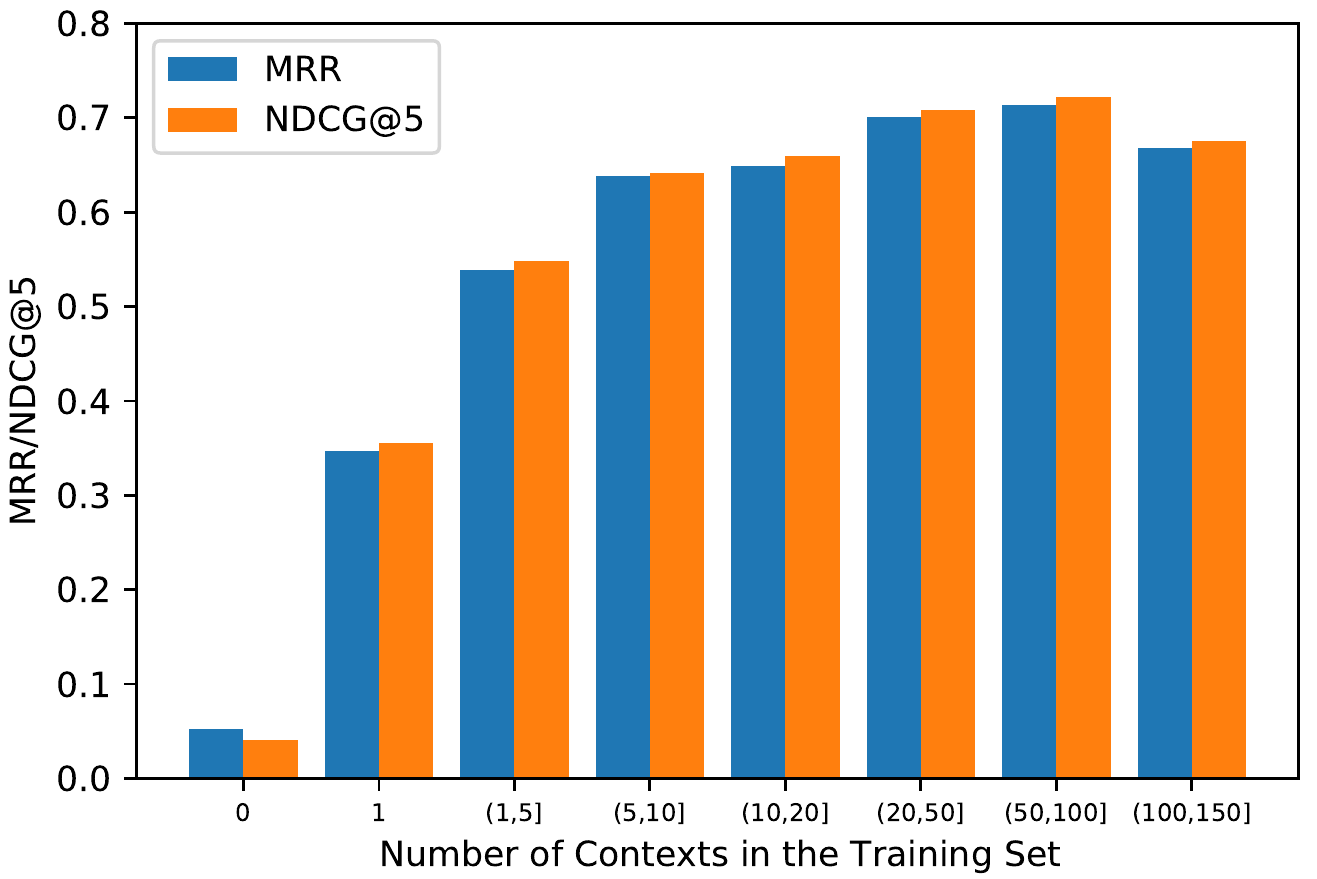}  
  \caption{Recommendation performance for quotes within different occurrence frequency ranges. The quote numbers in the ranges are 100, 843, 985, 437, 283, 225, 74 and 47, respectively.}
\label{fig:freq}
\end{figure}

\subsection{Quote Recommendation with Left Context Only}
\label{sec:left}
Following previous work \citep{tan2015learning,ahn2016quote,tan2018quoterec}, the evaluation experiments are mainly conducted in the setting where both the left and right contexts are given.
However, in practical terms, quote recommendation given the left context only might be more useful.
Therefore, we also conduct experiments in the setting where only the left context is given.
Table \ref{tab:left} shows the results. We can see that our method is still the best one on all three parts.
In addition, the performance of all methods decreases substantially, which indicates that both the left and right contexts provide important information for quote recommendation.

\subsection{Effect of Occurrence Frequency}
\label{sec:freq}
In this subsection, we investigate the effect of the gold quote's occurrence frequency on recommendation performance.
Figure \ref{fig:freq} shows MRR and NDCG@5 results for quotes that have different numbers of contexts in the training set of the standard Chinese part.

We observe that the occurrence frequency has great impact on quote recommendation performance.
Basically, increasing occurrences of quotes in the training set can increase recommendation performance, because we can learn better representations for the quotes with more adequate training.
But the most frequent quotes does not have the best performance, possibly because these quotes carry very rich semantics and can be cited in various contexts 
, which makes it very hard to correctly recommend them.
In addition, the performance for the unseen quotes is very limited.
It reflects the weakness of our model in the zero-shot situation, whose solution is left for future work.

\begin{table}[t!]
\centering
\resizebox{\linewidth}{!}{%
\begin{tabular}{@{}r|ccrc@{}}
\toprule
\#NS & MRR & NDCG & \multicolumn{1}{c}{$\tilde{R}$ /$\bar{R}$ /$\sigma_R$} & Recall@1/10/100 \\ \midrule
4 & 0.533 & 0.540 & \underline{2} / 161 / 412 & 47.48 / 63.23 / 77.68 \\
9 & 0.534 & 0.541 & \underline{2} / 148 / 381 & 47.50 / 63.97 / 78.83 \\
19 & {0.541} & {0.548} & \underline{2} / \underline{139} / 370 & \textbf{47.91} / \underline{64.97} / \underline{79.35} \\
29 & \textbf{0.545}	& \textbf{0.552} & \underline{2} / 174 / 434 & 47.06 / 63.58 / 76.92 \\
39 & 0.535 & 0.543 & \underline{2} / \underline{132} / \textbf{357} & 47.17 / \underline{64.97} / \underline{79.43} \\ \bottomrule
\end{tabular}%
}
\caption{Quote recommendation results with different negative sample numbers (\#NS).}
\label{tab:pnratio}
\end{table}

\subsection{Effect of Negative Sample Number}
\label{sec:ns}
In this subsection, we investigate the effect of the negative sample number (\#NS), a hyper-parameter of our method, on quote recommendation performance.
Table \ref{tab:pnratio} gives the results of different negative sample numbers on the validation set of the standard Chinese part of QuoteR.

We can see that increasing negative samples (from 4 to 19) can increase quote recommendation performance, which is because the quote encoder can be trained more sufficiently. 
However, when the negative samples continue increasing, the performance fluctuates or even decreases.
That is possibly because of the imbalance of positive and negative samples (there is only one positive sample, namely the gold quote), as explained in §\ref{sec:train}. 
Therefore, taking both performance and computation efficiency into consideration, we choose 19 as the final negative sample number.

\subsection{Human Evaluation}
\label{sec:human}
As mentioned in §\ref{sec:formalize}, there may be other quotes that are suitable for a query context besides the gold quote.
Hence, we conduct a human evaluation on the recommendation results of our method.
We randomly select 50 contexts from the validation set of the standard Chinese part and list the top 10 quotes recommended by our method for each context.
Then we ask annotators to make a binary suitability decision for the quotes.
Each quote is annotated by 3 native speakers and the final decision is made by voting.
For each context, we regard the suitable quote with the highest ranking as the gold quote, and re-evaluate the recommendation performance: NDCG@5=0.661, Recall@1/10=0.50/0.92.\footnote{Since we only annotate the top 10 results, there are no other available metrics than NDCG@5 and Recall@1/10.} 
In contrast, the original evaluation results among the 50 contexts are  NDCG@5=0.439 , Recall@1/10=0.36/0.64.
By comparison, we can conclude that the real performance of our method is substantially underestimated.
We also count the average number of suitable quotes among the top 10 quotes, which is 1.76. 

\begin{table}[t]
\centering
\resizebox{1\linewidth}{!}{%
\begin{tabular}{@{}cll@{}}
\toprule
Rank & \multicolumn{1}{c}{Quote} & \multicolumn{1}{c}{Score} \\ \hline
1 & {sufficient for the day is its own trouble} & 0.723 \\ \hline
2 & \textbf{sufficient unto the day is the evil thereof} & 0.124 \\ \hline
3 & \begin{tabular}[c]{@{}l@{}}you can never plan the future by the past\end{tabular} & 0.060 \\ \hline
4 & tomorrow will be a new day & 0.025 \\ \hline
5 & the darkest hour is just before the dawn & 0.008 \\ \bottomrule
\end{tabular}%
}
\caption{Top 5 results for the context in Figure \ref{fig:example}.}
\label{tab:case}
\end{table}

\subsection{Case Study}
We feed the context in Figure \ref{fig:example} into our model, and print the top 5 recommended quotes and their rank scores in Table \ref{tab:case}.
We find that the gold quote is ranked second, but the first one is actually another statement version of the gold quote and has exactly the same meaning.
In addition, the third and fourth quotes are also related to the context.
This case, together with more cases in Appendix \ref{sec:case}, can demonstrate the practical effectiveness and usefulness of our model.

\section{Conclusion and Future Work}
In this paper, we build a large and the first open dataset of quote recommendation for writing named QuoteR and conduct an extensive evaluation of existing quote recommendation methods on it.
We also propose a new model that achieves absolute outperformance over previous methods, and its effectiveness is proved by ablation studies.
In the future, we will try to improve our model in handling classical Chinese quotes by using a special classical Chinese pre-trained model to encode them. 
We will also consider boosting the performance of our model in the few-shot and zero-shot situations. 

\section*{Acknowledgements}
This work is supported by the National Key R\&D Program of China (No. 2020AAA0106502), Institute Guo Qiang at Tsinghua University, and International Innovation Center of Tsinghua University, Shanghai, China.
We also thank all the anonymous reviewers  for their valuable comments and suggestions.

\section*{Ethical Statements}
In this section, we discuss the ethical considerations of this paper from four perspectives.

\paragraph{Dataset and Human Evaluation}
In terms of our QuoteR dataset, all the quotes are collected from free and open quote repository websites.
Besides, all the contexts are extracted from open corpora, including free public domain e-books and other open corpora.
Therefore, there is no intellectual property problem for the dataset.
In addition, we conduct the human evaluation by a reputable data annotation company.
The annotators are fairly compensated by the company, based on the previous annotation tasks.
Further, we do not directly communicate with the annotators, so that their privacy is well preserved.
Finally, the dataset and the human evaluation are not sensitive and thus do not need to be approved by the institutional review board (IRB).

\paragraph{Application}
Quote recommendation is a practical task and our model can be put into service.
In actual use cases, users just need to input a query context and our model should output a list of candidate quotes that fit the given context.
All people may benefit from our model during writing.
If our model fails, some inappropriate quotes that cannot fit the query context would be output, but no one would be harmed.
There are indeed biases in the dataset we build.
Some quotes are very frequent while the others are not, as illustrated in §\ref{sec:freq}.
The infrequent quotes are less recommended and may cause the failure of our model in some cases.
In terms of misuse, to the best of our knowledge, such a quote recommendation model is hardly misused.
After the deployment of our model, the system would not collect data from users. 
It does not have any potential harm to vulnerable populations, either.

\paragraph{Energy Saving}
To save energy, we use the base version of BERT rather than larger pre-trained language models, although the larger ones would probably yield better performance.
Besides, as discussed in §\ref{sec:train}, we find that the simultaneous training of the context and quote encoders requires very big memory and computation resources, and thus we adopt the strategy of negative sampling in training.

\paragraph{Use of Identity Characteristics}
In this work, we do not use any demographic or identity characteristics information.

\bibliography{custom}
\bibliographystyle{acl_natbib}

\appendix
\begin{CJK}{UTF8}{gkai}

\section{More Statistics of QuoteR}
\label{sec:stat}
We count the numbers of quotes within different ranges of context-quote pair numbers, and the results are shown in Table \ref{tab:stat2-dataset}.
We can see the long tail, i.e., most quotes occur a few times while a small amount of quotes appear very frequently, which demonstrates the necessity of restricting the maximum number of contexts for a quote during the construction of QuoteR.

\begin{table}[!h]
\centering
\resizebox{\linewidth}{!}{
    \begin{tabular}{crrrrr}
    \toprule
	\multicolumn{6}{l}{\underline{English}}\\
  	\#Context &  [5,10] & (10,20] &(20,50] &(50,100] &(100,200] \\
	\#Quote &  2,994 & 1,456 & 1,233 & 257 & 168\\
  	\midrule
	\multicolumn{6}{l}{\underline{Standard Chinese}} \\
  	\#Context & [5,10] & (10,20] &(20,50] &(50,100] &(100,150] \\
	\#Quote & 2,207 & 371 & 272 & 79 & 75\\
	\midrule
	\multicolumn{6}{l}{\underline{Classical Chinese}} \\
  	\#Context & [3,10] & (10,20] &(20,50] &(50,100] &(100,150] \\
	\#Quote & 1,995 & 1,074 & 761 & 316 & 292\\
    \bottomrule
    \end{tabular}
}
\caption{The distribution of quotes within different occurrence frequency (the number of context-quote pairs) ranges of the three parts of QuoteR.}
\label{tab:stat2-dataset}
\end{table}

\section{More Case Studies}
\label{sec:case}
Table \ref{tab:case-mc}-\ref{tab:case-ac} show three quote recommendation cases for English, standard and classical Chinese, respectively.
\begin{itemize}
  \item For the standard Chinese case in Table \ref{tab:case-mc}, the gold quote is also ranked first properly. Moreover, the 2nd and 5th recommendations, which convey the meaning of ``change is the only constant thing in the world'', also fit the given context.
  \item For the English case in Table \ref{tab:case-en}, the gold quote is correctly ranked first. And the 2nd and 5th recommended quotes have the same meaning as the gold one, and thus suitable for the context as well.
  \item For the classical Chinese case in Table \ref{tab:case-ac}, the gold quote receives the highest rank score once again. And the 2nd recommended quote actually suits the context too. In addition, the 4th quote is also semantically related to the meaning of the context.
\end{itemize}

The three cases can demonstrate the effectiveness and practicability of our quote recommendation model.

\section{Reproducibility}
In this section, we report more experimental details to ensure the reproducibility of this paper.

All the experiments are conducted on a server that has 32 Intel(R) Xeon(R) Platinum 8163 @2.50GHz CPUs and 4 16-GB Nvidia Tesla V100 GPUs.
The operation system is Ubuntu 18.04.
We use Python v3.6.9 and PyTorch \citep{paszkepytorch} v1.7.1 to implement our model.
More details about the implementation, e.g., dependency libraries, can be found in the README file of the Software in the supplementary materials.

In addition, our models for English, standard Chinese and classical Chinese have about 308M, 308M and 329M parameters, respectively.
And the average training time is 7.5h, 26h and 29h, respectively.

\begin{table}[t!]
\centering
\resizebox{1.02\linewidth}{!}{%
\begin{tabular}{@{}cll@{}}
\toprule
Rank & \multicolumn{1}{c}{Quote} & \multicolumn{1}{c}{Score} \\ \hline
1 & \begin{tabular}[c]{@{}l@{}}\textbf{人不能两次踏进同一条河流} \\ \textbf{No man ever steps in the same river twice} \end{tabular} & 0.995 \\ 
\hline
2 & \begin{tabular}[c]{@{}l@{}}世界上唯一不变的就是变化 \\ The only constant in life is change \end{tabular} & 0.002 \\ 
\hline
3 & \begin{tabular}[c]{@{}l@{}}萧瑟秋风今又是，换了人间 \\ The autumn wind still sighs, but the world has changed\end{tabular} & 0.001 \\ 
\hline
4 & \begin{tabular}[c]{@{}l@{}}前途是光明的，道路是曲折的 \\ The road is tortuous, but the future is bright \end{tabular} & 0.001 \\ 
\hline
5 & \begin{tabular}[c]{@{}l@{}}只有变化是永恒的 \\ Change is the only constant \end{tabular} & 0.001 \\ 
\bottomrule
\end{tabular}%
}
\caption{A standard Chinese quote recommendation case. Top 5 recommended quotes (the gold quote is in boldface) are listed for the context: 从盘面上看，股票价格会呈现某种带漂移的无规则行走，涨跌无常，难以捉摸。[Quote]，这话放在投资领域也同样受用。事物是在不断变化的，历史数据只能起一定程度的参考作用。投资者想凭借历史数据准确预测未来几乎是不可能的。(\textit{The stock price shows some kind of irregular walk with drift, up and down unpredictably. The saying that [Quote] is also applicable to investment. Things are constantly changing, and historical data have limited reference value. It is almost impossible for investors to accurately predict the future based on historical data.})}
\label{tab:case-mc}
\end{table}

\begin{table}[t!]
\centering
\resizebox{.9\linewidth}{!}{%
\begin{tabular}{@{}cll@{}}
\toprule
Rank & \multicolumn{1}{c}{Quote} & \multicolumn{1}{c}{Score} \\ \hline
1 & \begin{tabular}[c]{@{}l@{}}\textbf{Truth is always strange} \end{tabular} & 0.984 \\ 
\hline
2 & \begin{tabular}[c]{@{}l@{}}{Truth is always stranger than fiction} \end{tabular} & 0.005 \\ 
\hline
3 & \begin{tabular}[c]{@{}l@{}} Truth is dangerous\end{tabular} & 0.002 \\ 
\hline
4 & \begin{tabular}[c]{@{}l@{}} Truth is subjectivity \end{tabular} & 0.001 \\ 
\hline
5 & \begin{tabular}[c]{@{}l@{}} Fact is stranger than fiction \end{tabular} & 0.001 \\ 

\bottomrule
\end{tabular}%
}
\caption{An English quote recommendation case. Top 5 recommended quotes (the gold quote is in boldface) are listed for the context: \textit{We've talked about some of the prophecies that have already come true in our cities from science fiction. What are some prophecies that have yet to come true? In a way, while sci-fi is fascinating, [Quote]. The transformation around surveillance is already mimicking a lot of the predictions in, say, minority report, which was very much emphasizing how surveillance and marketing were becoming completely tailored to the individual.}}
\label{tab:case-en}
\end{table}

\begin{table}[t!]
\centering
\resizebox{1.02\linewidth}{!}{%
\begin{tabular}{@{}cll@{}}

\toprule
Rank & \multicolumn{1}{c}{Quote} & \multicolumn{1}{c}{Score} \\ \hline
1 & \begin{tabular}[c]{@{}l@{}}\textbf{道不同，不相为谋} \\ \textbf{Persons walking different paths cannot work together} \end{tabular} & 0.412 \\ 
\hline
2 & \begin{tabular}[c]{@{}l@{}}话不投机半句多 \\ One word is too much for someone uncongenial \end{tabular} & 0.270 \\ 
\hline
3 & \begin{tabular}[c]{@{}l@{}}惺惺惜惺惺 \\ The wise appreciate one another\end{tabular} & 0.111 \\ 
\hline
4 & \begin{tabular}[c]{@{}l@{}}白头如新，倾盖如故 \\ You may know a little about old acquaintances and make \\ close friends with a stranger soon \end{tabular} & 0.033 \\ 
\hline
5 & \begin{tabular}[c]{@{}l@{}}近朱者赤，近墨者黑 \\ One takes the behavior of one's company\end{tabular} & 0.024 \\ 
\bottomrule
\end{tabular}%
}
\caption{A classical Chinese quote recommendation case. Top 5 recommended quotes (the gold quote is in boldface) are listed for the context: 我是少数群体中的一员，谈不上饱受社会不文明的欺压，却也受到主流文化对边缘群体的排斥。你不认可我，所谓[Quote]，我哪里还能跟你热情。相反，认可我的人就会享受我的回应，真诚也好，善良也好，温柔也好，我会把我好的一面展示给他们。(\textit{I am a member of a minority group, not suffering from the oppression of uncivilized people in society, but being ostracized by mainstream culture. If you don't approve of me, as the saying goes, [Quote], I can't be enthusiastic about you. In contrast, the persons who approve of me will enjoy my good side, including sincerity, goodness and gentleness.})}
\label{tab:case-ac}
\end{table}

%
%

\end{CJK}

\end{document}